\pdfoutput=1

\documentclass[sigconf,screen,authorversion,nonacm]{acmart}

\usepackage{fancyhdr}
\AtBeginDocument{%
    \addtolength{\footskip}{2.0\baselineskip}%
}

\usepackage{tikz}
\usetikzlibrary{arrows}
\tikzstyle{node} = [circle, draw, inner sep=2pt, fill=black]
\tikzstyle{wide_block} = [draw, align=left]
\tikzstyle{block} = [text width=4.25cm, minimum height=0.75cm, align=center]
\tikzstyle{small_block} = [text width=4.25cm, anchor=north]

\usepackage{makecell}
\usepackage{calc}

\AtBeginDocument{%
  }

\setcopyright{acmlicensed}
\copyrightyear{2025}
\acmYear{2025}
\acmDOI{10.48550/arXiv.2412.16641}

\acmConference[ACM FAccT]{ACM Conference on Fairness, Accountability, and Transparency}{2025}{Athens, Greece}
\acmISBN{978-1-4503-XXXX-X/18/06}

\begin{document}

\title{A Systems Thinking Approach to Algorithmic Fairness}

\author{Chris Lam}
\affiliation{%
  \institution{Epistamai}
  \city{Apex}
  \state{NC}
  \country{27502}}
\email{chris@epistam.ai}

\renewcommand{\shortauthors}{Lam}

\begin{abstract}
  Systems thinking provides us with a way to model the algorithmic fairness problem by allowing us to encode prior knowledge and assumptions about where we believe bias might exist in the data generating process. We can then encode these beliefs as a series of causal graphs, enabling us to link AI/ML systems to politics and the law. This allows us to combine techniques from machine learning, causal inference, and system dynamics in order to capture different emergent aspects of the fairness problem. We can use systems thinking to help policymakers on both sides of the political aisle to understand the complex trade-offs that exist from different types of fairness policies, providing a sociotechnical foundation for designing AI policy that is aligned to their political agendas and with society's shared democratic values.
\end{abstract}



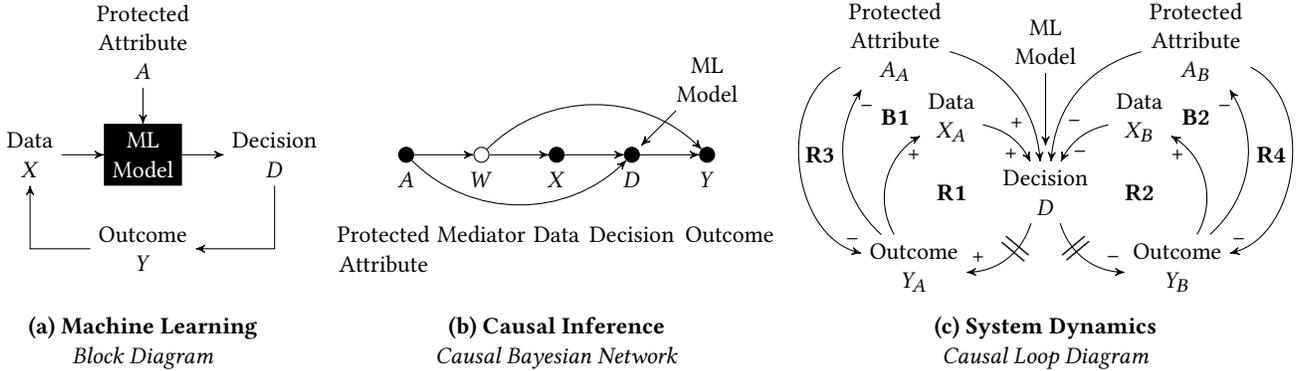
\begin{teaserfigure}
    \centering
    \begin{tikzpicture}[->, >=stealth', node distance=1cm, align=center]
        \node[node, label=below:$X$] (X) at (0,0) {};
        \node[node, label=below:$W$, fill=white] (W) [left of=X] {};
        \node[node, label=below:$A$] (A) [left of=W] {};
        \node[node, label=below:$D$] (D) [right of=X] {};
        \node[node, label=below:$Y$] (Y) [right of=D] {};
        \node (X2) [below of=X, yshift=-0.25cm] {Data\\};
        \node (W2) [left of=X2] {Mediator\\};
        \node (A2) [left of=W2, xshift=-0.3cm] {Protected\\Attribute};
        \node (D2) [right of=X2] {Decision\\};
        \node (Y2) [right of=D2, xshift=0.3cm] {Outcome\\};
        \node[above of=Y] (ML) {ML\\Model};
        \path
            (A) edge node {} (W)
            (W) edge node {} (X)
            (X) edge node {} (D)
            (D) edge node {} (Y)
            (W) edge[bend left=45] node {} (Y)
            (A) edge[bend right=45] node {} (D);
        \draw[->, shorten >=2pt] (ML) -- (D);
        \node[draw, align=center, fill=black, text=white] (MLB) at (-5.5,0) {ML\\Model};
        \node[left of=MLB, xshift=-0.5cm] (XB) {Data\\$X$};
        \node[right of=MLB, xshift=0.75cm] (DB) {Decision\\$D$};
        \node[above of=MLB, yshift=0.5cm] (AB) {Protected\\Attribute\\$A$};
        \node[below of=MLB, yshift=-0.25cm] (YB) {Outcome\\$Y$};
        \draw[->] (XB) -- (MLB);
        \draw[->] (MLB) -- (DB);
        \draw[->] (AB) -- (MLB);
        \draw (DB) |- (YB);
        \draw (YB) -| (XB);
        \node (DL) at (6.5,-0.5) {Decision\\$D$};
        \node (XA) at (5.25,0.5) {Data\\$X_A$};
        \node (XB) at (7.75,0.5) {Data\\$X_B$};
        \node (YA) at (4.75,-1.5) {Outcome\\$Y_A$};
        \node (YB) at (8.25,-1.5) {Outcome\\$Y_B$};
        \node (AA) at (4.5,1.5) {Protected\\Attribute\\$A_A$};
        \node (AB) at (8.5,1.5) {Protected\\Attribute\\$A_B$};
        \node (MLL) at (6.5,1.5) {ML\\Model};
        \path (DL) edge[bend left=35] node[very near end, above] {+} (YA);
        \path (DL) edge[bend right=35] node[very near end, above] {--} (YB);
        \path (YA) edge[bend left=45] node[near end, right] {+} (XA);
        \path (YB) edge[bend right=45] node[near end, left] {+} (XB);
        \path (XA) edge[bend left=25] node[very near end, left] {+} (DL);
        \path (XB) edge[bend right=25] node[very near end, right] {--} (DL);
        \path (YA) edge[bend left=40] node[very near end, right] {--} (AA);
        \path (YB) edge[bend right=40] node[very near end, left] {--} (AB);
        \path (AA) edge[bend left=35] node[near end, left] {+} (DL);
        \path (AB) edge[bend right=35] node[near end, right] {--} (DL);
        \path (AA) edge[bend right=70] node[very near end, right] {--} (YA);
        \path (AB) edge[bend left=70] node[very near end, left] {--} (YB);
        \draw[->, shorten >=6pt] (MLL) -- (DL);
        \node (R1) at (5.25,-0.5) {\textbf{R1}};
        \node (R2) at (7.75,-0.5) {\textbf{R2}};
        \node (B1) at (4.5,0.5) {\textbf{B1}};
        \node (B2) at (8.5,0.5) {\textbf{B2}};
        \node (R3) at (3.5,0) {\textbf{R3}};
        \node (R4) at (9.5,0) {\textbf{R4}};
        \node[rotate=50] at (6.1,-1.25) {\LARGE ||};
        \node[rotate=-50] at (6.9,-1.25) {\LARGE ||};
        \node (L1) at (0,-2.5) {\textbf{(b) Causal Inference}\\\textit{Causal Bayesian Network}};
        \node (L2) at (-5.5,-2.5) {\textbf{(a) Machine Learning}\\\textit{Block Diagram}};
        \node (L3) at (6.5,-2.5) {\textbf{(c) System Dynamics}\\\textit{Causal Loop Diagram}};
    \end{tikzpicture}
    \caption{Three representations of algorithmic fairness}
    \Description{A block diagram, a causal Bayesian network, and a causal loop diagram.}
    \label{fig:three_representations}
\end{teaserfigure}

\received{25 June 2025}

\maketitle

\section{Introduction}

For several years, AI researchers have struggled with how to build AI/ML systems that are fair and aligned to society's shared democratic values. Purely technical approaches that rely on correlation alone have fallen far short of what is needed to solve this difficult problem. Systems thinking offers a novel approach to the algorithmic fairness problem by allowing us to use causality to combine complex ideas from the social sciences and computer science into a comprehensive and integrated theory. By leveraging prior knowledge and assumptions about where we believe bias might exist in the data generating process, we can help policymakers on both sides of the aisle to design AI policy that is aligned to their political agendas.

To address the algorithmic fairness problem, we need to take a sociotechnical approach. The first part of this paper will briefly cover concepts from the social sciences to address the controversial question: Why do we believe disparities exist across protected groups like race or gender? This will give us insights into the political debate around fairness and why policymakers fundamentally disagree on how to close those disparities.

We will then turn our attention to the technical aspects of the fairness problem, which forms the bulk of this paper. We will use systems thinking to show how to combine analytical techniques from machine learning, causal inference, and system dynamics to model different facets of the fairness problem. Machine learning is necessary for building AI systems that can make fair predictions. Causal inference allows us to visualize different forms of bias, showing us what interventions are needed to make a machine learning model fair. System dynamics allows us to model bias from the data generating process, providing counterintuitive explanations as to why disparities continue to persist across protected groups. Finally, we will demonstrate how these three techniques can form the basis of a new causal hierarchy for modeling algorithmic fairness.

\section{Related work}

The field of algorithmic fairness is a relatively new field that has drawn significant attention since the release of a ProPublica article that claimed racial bias in COMPAS, a software tool that is widely used in the US court system to predict recidivism\cite{angwin}. Since then, there have been multiple attempts to align the legal definitions of discrimination (e.g. disparate treatment and disparate impact in the US or direct discrimination and indirect discrimination in the EU) with their equivalent mathematical definitions. But these approaches, which have oftentimes relied on statistical correlation, have not been successful. Narayanan for example identified nearly two dozen fairness metrics, none of which could fully align to the legal definitions of discrimination\cite{narayanan}.

Pearl has argued for a few decades that the discrimination problem is fundamentally a causality problem\cite{10.5555/2074022.2074073}. He has argued that fairness can be modeled using causal Bayesian networks (CBNs) as a series of direct, indirect, and spurious effects. This work was later expanded upon by Plecko and Bareinboim, who proposed the idea of a standard fairness model\cite{plecko2022causalfairnessanalysis}. They argued that there exists a causal template through which entire classes of fairness models can be derived using a small number of modeling assumptions. Kusner et al. argued for the use of counterfactuals in modeling fairness\cite{NIPS2017_a486cd07}. Chiappa and Isaac showed that certain paths within a CBN could be labeled as fair or unfair\cite{Chiappa_2019}. Further attempts at building causal fairness models were made by Nilforoshan et al.\cite{nilforoshan2022causalconceptionsfairnessconsequences} and Kilbertus et al.\cite{10.5555/3294771.3294834}. Barocas et al. devoted a chapter in their textbook on fairness and machine learning to discuss how causality might be used to model fairness\cite{barocas-hardt-narayanan}.

However, all of their approaches were limited due to their inability to construct a valid causal model of the problem. This is because their approaches did not fully take into account the social and political aspects of the fairness problem. Society has different ideological beliefs about what is fair and unfair based on different sets of assumptions about where bias is within the data generating process. Our paper addresses this issue by encoding society's different mental models of bias into causal models using first principles. By taking into account a diversity of perspectives, we can understand the normative aspects of the fairness problem. This is necessary for linking machine learning to politics and ultimately to the law.

Another key difference is that their causal approaches relied solely on causal inference, which has limitations in its ability to model the dynamics of the data generating process. Our approach leverages systems thinking and system dynamics to gain a fuller understanding of the data generating process through the modeling of feedback loops. Martin has also explored modeling AI bias using community-based system dynamics\cite{martin, martin2}. Furthermore, we link supervised machine learning to causal inference by recognizing that a CBN can be used as a higher level abstraction of a black box ML model. By bringing supervised machine learning, causal inference, and systems thinking together, we can model different aspects of the algorithmic fairness problem.

\section{Towards a unified theory of fairness} \label{theory}

To build fair AI/ML systems, we begin from first principles with the strong assumption that all humans have free will\cite{carus}. This forms the basis of moral responsibility\cite{sep-moral-responsibility}. If we humans have free will, then we have the agency to make our own decisions. These decisions will lead to certain outcomes, and we may be held morally responsible for those outcomes.

Most of the time, we humans make decisions that only affect our own selves. We can be held personally responsible for the decisions that we make for ourselves. But sometimes, we humans make decisions that affect other people. This is especially true for humans who are in positions of great power. We may be held socially responsible for the decisions that we make for others.

Sometimes the decisions that we make will lead to good outcomes for other people. In these cases, they might assign us praise and reward. On the other hand, sometimes the decisions that we make will lead to bad outcomes for other people. In these cases, they might assign us blame and punishment. This concept is known as moral desert\cite{10.1093/acprof:oso/9780199895595.003.0001}. We humans have a natural tendency to accept praise and reward, and to reject blame and punishment.

But sociologists would note that we are never fully free to make our own decisions. This is because social structures can influence our agency\cite{barker}. Social structures define our relationships with the various institutions of society and may include our economic systems, legal frameworks, cultural norms, and political institutions. Agency defines our ability to act independently and to freely make our own decisions. Social structures place a limit on the the choices and opportunities that are available to us humans. Those who are economically disadvantaged will encounter greater limits to their agency due to social structures compared to those who are economically advantaged.

To better understand human behavior, we turn to a concept in psychology called the locus of control. This was coined by Rotter and refers to the degree of belief that people have control over the outcomes of events in their own lives\cite{rotter}. Humans with an internal locus of control believe that they are in control of their own lives, while those with an external locus of control believe that their lives are controlled by outside factors. These concepts might also be related to whether someone adopts a growth mindset\cite{dweck} or a victimhood mindset\cite{kaufman}.

Psychologists provide another key concept for fairness called attribution theory, which is used by humans to interpret and assign causes for events. First proposed by Heider\cite{heider} and later expanded by Kelley\cite{kelley} and Weiner\cite{weiner}, attribution theory examines whether attributions should be assigned to internal (dispositional) attributions or external (situational) attributions. The internal attributions may be due to behavior or personal traits, while the external attributions may be due to situational or environmental factors. This is essential for determining whether we believe an outcome is biased and therefore unfair.

For example, let's say that a student did poorly on an exam. An internal attribution might argue that the student was lazy and should have spent more time studying. In this case, the student is to blame for the poor results and thus the test was unbiased and fair. An external attribution might argue that the test was too hard or that the material was not taught well. In this case, the professor is to blame for the poor results and thus the test was biased and unfair. 

To determine whether an outcome is biased and unfair, all we need to know is whether we would attribute a bad outcome as being internal or external. This makes algorithmic fairness a type of credit assignment problem. But since we usually focus on bad outcomes, this oftentimes becomes a blame assignment problem. That being said, identifying attribution is much harder in practice, as we will show later when we model fairness using system dynamics. Both internal and external factors may influence and reinforce each other, leading to vicious feedback loops. This becomes a particularly difficult problem when trying to explain disparities across protected groups like race, which we describe in greater detail in Section \ref{racial_disparities}.

This causality dilemma between attributing disparities to internal or external factors ultimately leads to the controversial and heated political debate over how best to reduce disparities across protected groups\cite{sowell_discrimination, macdonald, mills, kendi, loury}. Those on the political left are more likely to attribute disparities as being due to external factors, which provides the moral justification for strong government interventions like affirmative action or DEI. The left also tends to focus more on legal forms of discrimination like disparate impact or indirect discrimination. Those on the political right are more likely to attribute disparities as being due to internal factors, which provides the moral justification for limited government intervention like fairness through unawareness. The right also tends to focus more on legal forms of discrimination like disparate treatment or direct discrimination. This ideological debate is fundamental to understanding algorithmic fairness, but has been largely missing from the academic literature.

We will use systems thinking to encode these different ideological beliefs into a series of causal graphs  as shown in Figure \ref{fig:three_representations}. This will help us to understand what types of interventions policymakers on both sides of the political aisle might want to make to ensure that a machine learning model is fair. Finally, this could give us insights into how to build machine learning models that are compliant with antidiscrimination laws.

\section{Modeling fairness as a machine learning problem}

Before we build a causal representation of the fairness problem, we will start by modeling fairness as a supervised machine learning problem using a simple block diagram as shown in Figure \ref{fig:three_representations}a. In the center is a black box machine learning model that receives data $X$ as input and returns decision $D$ as output. The decision $D$ causes an outcome $Y$ that could potentially be recorded as future data $X$, thus creating a feedback loop. The key question that we need to address is whether the protected attribute $A$ should be used in a machine learning model and if so, how?

\section{Modeling fairness as a causal inference problem}

We can transform the block diagram from Figure \ref{fig:three_representations}a into a causal Bayesian network (CBN) as shown in Figure \ref{fig:three_representations}b. CBNs were pioneered by Pearl\cite{pearl_causality, pearl_why, pearl_inference} and allow us to symbolically encode knowledge and assumptions about causal effects between variables, which are shown in this graph as nodes. The first node of the CBN represents a protected attribute $A$, which has a direct effect on a mediator $W$ ($A \rightarrow W$). The mediator $W$ has a direct effect on an outcome $Y$ ($W \rightarrow Y$). We assume that the mediator $W$ completely explains away the causal effect between the protected attribute $A$ and the outcome $Y$, resulting in conditional independence ($Y \perp\!\!\!\perp A | W$). However, the mediator $W$ might be considered biased by some people if they attribute the disparities across protected groups as being external. The mediator $W$ is filled white because it is latent, but it can still be partially measured as data $X$ ($W \rightarrow X$). Data $X$ is then fed into a machine learning model that is embedded inside the node for decision $D$ ($X \rightarrow D$). Finally, the decision $D$ also has a direct effect on an outcome $Y$ ($D \rightarrow Y$). CBNs are represented by directed acyclic graphs (DAGs), which means that we cannot use them to model feedback loops. Examples of what each of these variables could represent are shown in Table \ref{tab:applications}. For each of these applications, the protected attribute $A$ could represent race or gender for example.

\begin{table*}[htbp!]
    \centering
    \caption{Examples of high stakes applications for algorithmic fairness}
    \begin{tabular}{c c c c c c}
        \toprule
        \textbf{Application} & \textbf{Mediator $W$} & \textbf{Data $X$} & \textbf{Decision $D$} & \textbf{Outcome $Y$}\\
        \midrule
        Credit & Creditworthiness & Income, credit history & Deny loan? & Loan default?\\
        Insurance & Insurability & Demographics, claims history & Deny policy? & Claim filing?\\
        Housing & Home value & Location, size, condition & Home valuation? & Final sales price?\\
        Employment & Qualifications & Experience, education & Screen resume? & Turnover?\\
        Education & Merit & Grades, test scores & Reject applicant? & Graduation?\\
        Research & Merit & Quality, clarity, originality & Reject paper? & No. of citations?\\
        \bottomrule
    \end{tabular}
    \Description{A table defining examples of variables for several high stakes applications.}
    \label{tab:applications}
\end{table*}

\begin{figure*}[htbp!]
    \begin{tabular}{>{\centering\arraybackslash} m{3.5cm} >{\centering\arraybackslash} m{4.6cm} >{\centering\arraybackslash} m{4.6cm} >{\centering\arraybackslash} m{3.5cm}}
        \makecell{\textbf{(a) Fairness through} \\\textbf{supremacism}} &
        \begin{tikzpicture}[->, >=stealth', node distance=1cm]
    	\node[node, label=below:$A$] (A) {};
    	\node[node, label=below:$W$, color=blue, fill=white] (W) [right of=A] {};
    	\node[node, label=below:$X$, color=blue] (X) [right of=W] {};
    	\node[node, label=below:$D$] (D) [right of=X] {};
    	\node[node, label=below:$Y$, color=blue] (Y) [right of=D] {};
    	\path
    		(A) edge[color=blue] node {} (W)
    		(W) edge[color=blue] node {} (X)
    		(X) edge[color=blue] node {} (D)
    		(D) edge node {} (Y)
    		(W) edge[bend left=45, color=blue] node {} (Y)
                (A) edge[bend right=45, color=red] node {} (D);
        \end{tikzpicture} &
        \begin{tikzpicture}[->, >=stealth', node distance=1cm]
    	\node[node, label=below:$A$] (A) {};
    	\node[node, label=below:$W$, fill=white] (W) [right of=A] {};
    	\node[node, label=below:$X$] (X) [right of=W] {};
    	\node[node, label=below:$D$, color=red] (D) [right of=X] {};
    	\node[node, label=below:$Y$, color=red] (Y) [right of=D] {};
    	\path
    		(A) edge node {} (W)
    		(W) edge node {} (X)
    		(X) edge node {} (D)
    		(D) edge[color=red] node {} (Y)
    		(W) edge[bend left=45] node {} (Y)
                (A) edge[bend right=45, color=red] node {} (D);
        \end{tikzpicture} &
        \makecell{\textbf{(b) Overt} \\\textbf{discrimination}} \\
        \makecell{\textbf{(c) Fairness through} \\\textbf{unawareness}} &
        \begin{tikzpicture}[->, >=stealth', node distance=1cm]
    	\node[node, label=below:$A$] (A) {};
    	\node[node, label=below:$W$, fill=white] (W) [right of=A] {};
    	\node[node, label=below:$X$] (X) [right of=W] {};
    	\node[node, label=below:$D$] (D) [right of=X] {};
    	\node[node, label=below:$Y$] (Y) [right of=D] {};
    	\path
    		(A) edge node {} (W)
    		(W) edge node {} (X)
    		(X) edge node {} (D)
    		(D) edge node {} (Y)
    		(W) edge[bend left=45] node {} (Y)
                (A) edge[bend right=45, color=white] node {} (D);
        \end{tikzpicture} &
        \begin{tikzpicture}[->, >=stealth', node distance=1cm]
    	\node[node, label=below:$A$] (A) {};
    	\node[node, label=below:$W$, color=red, fill=white] (W) [right of=A] {};
    	\node[node, label=below:$X$, color=red] (X) [right of=W] {};
    	\node[node, label=below:$D$, color=red] (D) [right of=X] {};
    	\node[node, label=below:$Y$, color=red] (Y) [right of=D] {};
    	\path
    		(A) edge[color=red] node {} (W)
    		(W) edge[color=red] node {} (X)
    		(X) edge[color=red] node {} (D)
    		(D) edge[color=red] node {} (Y)
    		(W) edge[bend left=45, color=red] node {} (Y)
                (A) edge[bend right=45, color=white] node {} (D);
        \end{tikzpicture} &
        \makecell{\textbf{(d) Covert} \\\textbf{discrimination}} \\
        \makecell{\textbf{(e) Fairness through} \\\textbf{affirmative action}} &
        \begin{tikzpicture}[->, >=stealth', node distance=1cm]
    	\node[node, label=below:$A$] (A) {};
    	\node[node, label=below:$W$, color=red, fill=white] (W) [right of=A] {};
    	\node[node, label=below:$X$, color=red] (X) [right of=W] {};
    	\node[node, label=below:$D$] (D) [right of=X] {};
    	\node[node, label=below:$Y$, color=red] (Y) [right of=D] {};
    	\path
    		(A) edge[color=red] node {} (W)
    		(W) edge[color=red] node {} (X)
    		(X) edge[color=red] node {} (D)
    		(D) edge node {} (Y)
    		(W) edge[bend left=45, color=red] node {} (Y)
                (A) edge[bend right=45, color=blue] node {} (D);
        \end{tikzpicture} &
        \begin{tikzpicture}[->, >=stealth', node distance=1cm]
    	\node[node, label=below:$A$] (A) {};
    	\node[node, label=below:$W$, fill=white] (W) [right of=A] {};
    	\node[node, label=below:$X$] (X) [right of=W] {};
    	\node[node, label=below:$D$, color=blue] (D) [right of=X] {};
    	\node[node, label=below:$Y$, color=blue] (Y) [right of=D] {};
    	\path
    		(A) edge node {} (W)
    		(W) edge node {} (X)
    		(X) edge node {} (D)
    		(D) edge[color=blue] node {} (Y)
    		(W) edge[bend left=45] node {} (Y)
                (A) edge[bend right=45, color=blue] node {} (D);
        \end{tikzpicture} &
        \makecell{\textbf{(f) Reverse} \\\textbf{discrimination}} \\
        \makecell{\textbf{(g) Fairness through} \\\textbf{lottery}} &
        \begin{tikzpicture}[->, >=stealth', node distance=1cm]
    	\node[node, label=below:$A$] (A) {};
    	\node[node, label=below:$W$, color=red, fill=white] (W) [right of=A] {};
    	\node[node, label=below:$X$, color=red] (X) [right of=W] {};
    	\node[node, label=below:$D$] (D) [right of=X] {};
    	\node[node, label=below:$Y$, color=red] (Y) [right of=D] {};
    	\path
    		(A) edge[color=red] node {} (W)
    		(W) edge[color=red] node {} (X)
    		(X) edge[color=white] node {} (D)
    		(D) edge node {} (Y)
    		(W) edge[bend left=45, color=red] node {} (Y)
                (A) edge[bend right=45, color=white] node {} (D);
        \end{tikzpicture} &
        \begin{tikzpicture}[->, >=stealth', node distance=1cm]
    	\node[node, label=below:$A$] (A) {};
    	\node[node, label=below:$W$, fill=white] (W) [right of=A] {};
    	\node[node, label=below:$X$] (X) [right of=W] {};
    	\node[node, label=below:$D$, color=blue] (D) [right of=X] {};
    	\node[node, label=below:$Y$, color=blue] (Y) [right of=D] {};
    	\path
    		(A) edge node {} (W)
    		(W) edge node {} (X)
    		(X) edge[color=white] node {} (D)
    		(D) edge[color=blue] node {} (Y)
    		(W) edge[bend left=45] node {} (Y)
                (A) edge[bend right=45, color=white] node {} (D);
        \end{tikzpicture} &
        \makecell{\textbf{(h) ``No''} \\\textbf{discrimination}} \\
    \end{tabular}
    \caption{Modeling bias, fairness, and discrimination using causal Bayesian networks (CBNs)}
    \Description{A series of causal Bayesian networks modeling different forms of bias, fairness, and discrimination.}
    \label{fig:cbn}
\end{figure*}
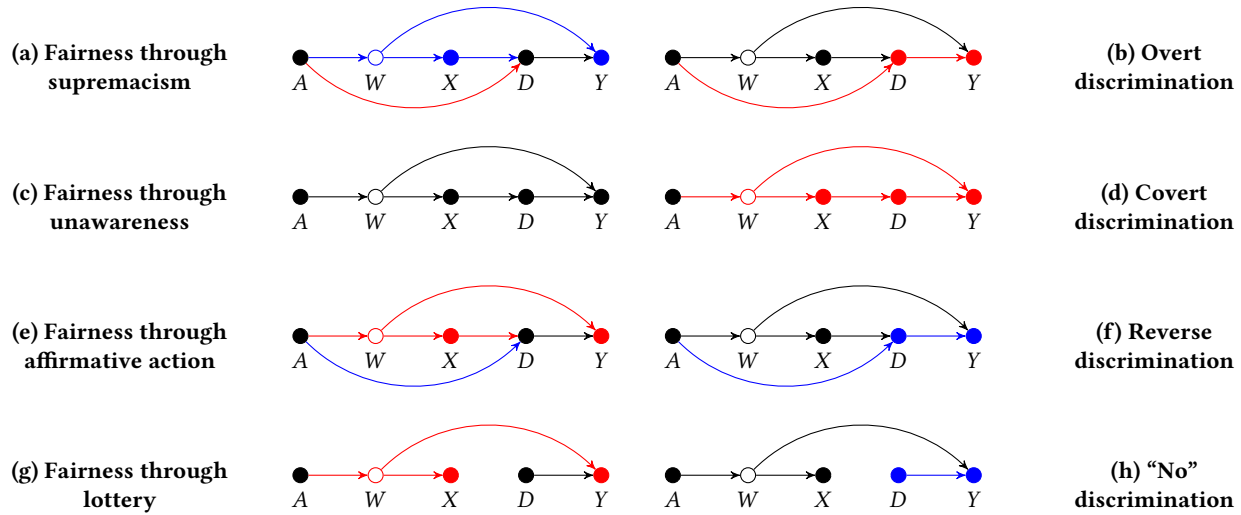

Figure \ref{fig:cbn} shows a series of CBNs modeling different forms of bias, fairness, and discrimination. Each CBN captures a different way of framing the fairness problem based on an individual's prior beliefs about where bias might exist in a model. On the left, we have four different forms of fairness, which we will later show can be mapped to different points on the political spectrum. On the right, we have four different forms of discrimination, which can be mapped directly to antidiscrimination law.

The first form of fairness that we will look at is \textbf{fairness through unawareness} (Figure \ref{fig:cbn}c). This is the simplest form of fairness. We begin with the belief that the data is fundamentally unbiased (internal attribution), which is why all of the nodes in the graph are colored black. In this form of fairness, there is no edge between the protected attribute $A$ and the decision $D$. Therefore, the model is blind to an individual's membership to a protected group $A$ and therefore cannot treat any individual differently on that basis. We would say that the protected attribute $A$ has no direct effect on the decision $D$, making the decision fair.

The advantage of fairness through unawareness is that it avoids \textbf{overt discrimination} (Figure \ref{fig:cbn}b). If a model uses knowledge of an individual being a member of a disadvantaged group against them in its decision $D$, then it would clearly cause disparate treatment or direct discrimination. We show this as a form of bias where the edge from the protected attribute $A$ to decision $D$ is colored red (\textcolor{red}{$A \rightarrow D$}). This leads to a biased decision $D$, which in turn causes a biased outcome $Y$ (\textcolor{red}{$D \rightarrow Y$}). Note that in these CBNs, bias is represented by colored nodes and flows across the entire graph. We would say that the protected attribute $A$ has a negative direct effect on the decision $D$ (\textcolor{red}{$A \rightarrow D$}), leading to harmful discrimination.

But if we instead believe that the data is fundamentally biased (external attribution), then fairness through unawareness could lead to \textbf{covert discrimination} (Figure \ref{fig:cbn}d). In this scenario, there is bias between the protected attribute $A$ and the mediator $W$ (\textcolor{red}{$A \rightarrow W$}), which flows across the entire graph and causes all of the remaining nodes in the graph to be colored red. This bias can be attributed to structural or systemic forms of discrimination that unfairly disadvantages one group over another. For example, if a disadvantaged group was overtly discriminated against in the past, then the present data $X$ may still reflect that negative bias. By removing the edge from the protected attribute $A$ to the decision $D$, it could reinforce one group's prior success over another. Covert discrimination could form the basis of a disparate impact or indirect discrimination claim. We would say that the protected attribute $A$ has a negative indirect effect on the decision $D$ (\textcolor{red}{$A \rightarrow W \rightarrow X \rightarrow D$}), leading to harmful discrimination.

To correct for covert discrimination, one might choose \textbf{fairness through affirmative action} (Figure \ref{fig:cbn}e). In this form of fairness, the protected attribute $A$ is used in the decision $D$ (\textcolor{blue}{$A \rightarrow D$}) to correct for bias from the data $X$. The intervention by the protected attribute $A$ is colored blue, which offsets the bias from the data $X$ that has been colored red. As a result, the final decision $D$ is colored black to indicate that its bias has been corrected. Interestingly, the outcome $Y$ is still colored red as the protected attribute $A$ does not have a direct effect on the outcome $Y$. The protected attribute $A$ instead has an indirect effect on the outcome $Y$ through the decision $D$, but this tends to be a weaker effect. This intervention does not directly fix the underlying biases in the mediator $W$, which has a strong direct effect on the outcome $Y$ (\textcolor{red}{$W \rightarrow Y$}). We would say that the protected attribute $A$ has a negative indirect effect on the decision $D$ (\textcolor{red}{$A \rightarrow W \rightarrow X \rightarrow D$)} and a positive direct effect on the decision $D$ (\textcolor{blue}{$A \rightarrow D$}), thus canceling out the biases and making the decision fair.

But if we disagree with the belief that the data is fundamentally biased (internal attribution), then fairness through affirmative action could be perceived as a form of \textbf{reverse discrimination} (Figure \ref{fig:cbn}f). The intervention of the protected attribute $A$ on the decision $D$ (\textcolor{blue}{$A \rightarrow D$}) could lead to individuals receiving a more positive decision $D$ but a more negative outcome $Y$. For example, a machine learning model may predict that an individual from a historically disadvantaged group may end up with a negative outcome $Y$. However, this would lead to a greater disparity in decisions across protected groups. So the model may instead use knowledge of that individual's membership in a protected group $A$ to render a more positive decision $D$. But this positive decision $D$ could lead to a worse outcome $Y$ for the individual than if that individual had just received a negative decision $D$ in the first place. An example might be approving a minority borrower for a loan even if that borrower is not creditworthy, which could result in a loan default. We would say that the protected attribute $A$ has a positive direct effect on the decision $D$ (\textcolor{blue}{$A \rightarrow D$}), leading to harmful discrimination.

We now move on to the more extreme forms of fairness. Reverse discrimination could lead to \textbf{fairness through supremacism} (Figure \ref{fig:cbn}a). An advantaged group may use this approach to actively suppress a disadvantaged group (e.g. white supremacism). This could happen if an advantaged group believes that a disadvantaged group has been unfairly receiving preferential treatment. For example, if a disadvantaged group has benefited from reverse discrimination in the past, then the present data $X$ may still reflect that positive bias and is colored blue. The protected attribute $A$ is then used to correct for that positive bias in the decision $D$ (\textcolor{red}{$A \rightarrow D$}) and is colored red. This might explain what happened when hiring managers racially discriminated against job applicants with Black sounding names compared to those with White sounding names even through the resumes were the same, as shown by Bertrand and Mullainathan\cite{NBERw9873}. We would say that the protected attribute $A$ has a positive indirect effect on the decision $D$ (\textcolor{blue}{$A \rightarrow W \rightarrow X \rightarrow D$}) and a negative direct effect on the decision $D$ (\textcolor{red}{$A \rightarrow D$}), thus canceling out the biases and making the decision fair.

Another extreme form of fairness is \textbf{fairness through lottery} (Figure \ref{fig:cbn}g). If we believe that the data is fundamentally biased, an alternative approach to removing bias in the decision $D$ is to remove the edge from data $X$ to decision $D$. The advantage of this approach to fairness is that it provides stronger guarantees of equalizing decision rates across groups. This is particularly effective for dealing with intersectionality issues, ensuring that every potential intersection of a protected class (e.g. race, gender, and religion) receives the same decision rates. Individuals cannot change the decision by altering their own data $X$. The decision $D$ could be assigned randomly or there could be a fixed rule where everyone receives a positive decision $D$ or a negative decision $D$. We would say that the protected attribute $A$ has no direct effect on the decision $D$ and its indirect effects have been canceled out, making the decision fair.

But fairness through lottery could be interpreted as \textbf{``no'' discrimination} (Figure \ref{fig:cbn}h). Since data $X$ has no effect on the decision $D$, then the machine learning model can literally not discriminate any differences between individuals. In doing so, it cannot cause illegal discrimination either. But this could still be interpreted as a decision that is positively biased in favor of a disadvantaged group. We would say that the protected attribute $A$ has a positive indirect effect on the decision $D$ (\textcolor{blue}{$A \rightarrow W \rightarrow X \not\rightarrow D$}), leading to harmful discrimination.

Note that this is not an exhaustive list of all forms of fairness and discrimination. For example, we could have a form of covert discrimination where the edges of Figure \ref{fig:cbn}d are all colored in blue instead of red.

\subsection{Mapping fairness to politics}

Table \ref{tab:politics} shows how each form of fairness can be mapped to different positions on the political spectrum. Each position is associated with its own fairness criterion and fairness properties.

\begin{table}[htbp!]
    \caption{Mapping fairness type to the political spectrum}
    \begin{tabular}{c c c c c}
        \toprule
        \makecell{\textbf{Political}\\\textbf{worldview}} & \makecell{\textbf{Far-}\\\textbf{left}} & \makecell{\textbf{Left-}\\\textbf{wing}} & \makecell{\textbf{Right-}\\\textbf{wing}} & \makecell{\textbf{Far-}\\\textbf{right}}\\
        \midrule
        \textbf{Type} & Lottery & \makecell{Affirmative\\action} & \makecell{Unaware-\\ness} & \makecell{Suprem-\\acism}\\
        \makecell{\textbf{Fairness}\\\textbf{criterion}} & $D \perp\!\!\!\perp A, W$ & $D \perp\!\!\!\perp A$ & $D \perp\!\!\!\perp A | W$ & $D\:\neg\!\!\perp\!\!\!\perp A$\\
        \makecell{\textbf{Individual}\\\textbf{fairness}} & Yes & No & Yes & No\\
        \makecell{\textbf{Group}\\\textbf{fairness}} & Yes & Yes & No & No\\
        \makecell{\textbf{Pearl's}\\\textbf{law}} & & First & Second & \\
        \bottomrule
    \end{tabular}
    \Description{A table describing different forms of fairness with points on the political spectrum.}
    \label{tab:politics}
\end{table}

Right-wing notions of fairness rely on fairness through unawareness, which is associated with the criterion $D \perp\!\!\!\perp A | W$. That is, the decision $D$ is conditionally independent of the protected attribute $A$ given some mediator $W$. For example, a college admissions decision is independent of race given an applicant's merit. This leads to individual fairness, where similar individuals are treated similarly regardless of protected group membership\cite{castelnovo, 10.1145/3194770.3194776, 10.1145/3457607, binns2019apparentconflictindividualgroup}. This form of fairness relies on Pearl's second fundamental law of causal inference: conditional independence.

Left-wing notions of fairness rely on affirmative action, which is associated with the criterion $D \perp\!\!\!\perp A$. That is, the decision $D$ is independent of the protected attribute $A$, which is achieved by using the protected attribute $A$ to correct for biases in the data $X$ that is used in the decision $D$. We imagine what the applicant's decision would have been had they been a member of a counterfactual class. We then perform an intervention to obtain the same decision. For example, a college admissions decision would affirmatively use race to correct for deficiencies in the applicant's merit due to systemic racism. This leads to group fairness, where different groups receive more similar aggregate decision rates\cite{castelnovo, 10.1145/3194770.3194776, 10.1145/3457607, binns2019apparentconflictindividualgroup}. This form of fairness relies on Pearl's first fundamental law of causal inference: counterfactuals and interventions.

We now examine the two extreme forms of fairness. Far-right notions of fairness rely on fairness through supremacism, which is associated with the criterion $D\:\neg\!\!\perp\!\!\!\perp A$. In this case, individual and group notions of fairness are both violated. Far-left notions of fairness rely on fairness through lottery, which is associated with the criterion $D \perp\!\!\!\perp A, W$. In this case, individual and group notions of fairness are both satisfied.

\subsection{Mapping discrimination to antidiscrimination law}

Table \ref{tab:law} shows how each form of discrimination can be mapped to antidiscrimination law. There are two key legal concepts in antidiscrimination law. In the US, these are called disparate treatment and disparate impact. In the EU, these are called direct discrimination and indirect discrimination. Disparate treatment or direct discrimination occurs when an individual or group is treated differently based on a protected characteristic like race or gender. Disparate impact or indirect discrimination occurs when a seemingly neutral policy has a disproportionately adverse impact on a protected group.

\begin{table}[htbp!]
    \caption{Mapping discrimination type to antidiscrimination law}
    \begin{tabular}{c c c}
        \toprule
        \makecell{\textbf{US}\\\textbf{Type}} & \makecell{\textbf{Disparate}\\\textbf{treatment}} & \makecell{\textbf{Disparate}\\\textbf{impact}}\\
        \midrule
        \makecell{\textbf{EU}\\\textbf{Type}} & \makecell{\textbf{Direct}\\\textbf{discrimination}} & \makecell{\textbf{Indirect}\\\textbf{discrimination}}\\
        \midrule
        \textbf{Overt} & X & \\
        \textbf{Covert} & & X \\
        \textbf{Reverse} & X & \\
        \textbf{``No''} & & X \\
        \bottomrule
    \end{tabular}
    \Description{A table describing how to map causal forms of discrimination with legal forms of discrimination.}
    \label{tab:law}
\end{table}

When a protected attribute $A$ has a direct effect on a decision $D$ ($A \rightarrow D$), we might consider this as a form of disparate treatment or direct discrimination. This could map to overt discrimination (Figure \ref{fig:cbn}b) and reverse discrimination (Figure \ref{fig:cbn}f). To avoid disparate treatment or direct discrimination, the protected attribute $A$ should have no direct effect on the decision $D$.

Related to disparate treatment is the concept of redlining, where a variable that is used in a decision $D$ acts as a proxy for a protected attribute $A$. For example, lenders have historically used zip code as a proxy for race as a basis for denying loans. We avoid redlining problems in this paper by only using data $X$ that is a direct measure of the mediator $W$. But a deeper discussion for how to prevent redlining using CBNs is discussed by Lam\cite{lam2024debiasingalternativedatacredit}. He shows that certain types of alternative data can cause redlining because the protected attribute $A$ may act as a hidden confounder, leading to that data having a spurious effect on a decision $D$.

When a protected attribute $A$ has an indirect effect on a decision $D$, we might consider this as a form of disparate impact or indirect discrimination. This could map to covert discrimination (Figure \ref{fig:cbn}d) and ``no'' discrimination (Figure \ref{fig:cbn}h). Even though the protected attribute $A$ has no direct effect on a decision $D$ (making the policy neutral), the inclusion of biased data or exclusion of unbiased data could cause the protected attribute $A$ to have an indirect effect on the decision $D$ that adversely affects a protected group.

For disparate impact in the US, there is a three-step test\cite{justice}. The first step is to establish that a neutral policy has an adverse disparate impact on a protected group. This would involve testing whether there is a disparity in decisions $D$ across the protected class $A$. The second step is to establish the business necessity of that policy. This would involve arguing that the data $X$ being used has some causal relationship to the mediator $W$. The final step is to identify a less discriminatory alternative model, which would involve performing a search for a counterfactual model.

CBNs provide an analytical framework to reason about whether a model is causing illegal discrimination. By being able to explicitly model the different forms of discrimination, CBNs provide a common language that businesses and regulators can use to ensure that machine learning models are compliant with laws and regulations.

\subsection{Normative effects of fairness and discrimination} \label{normative}

CBNs are a very effective way to model the normative effects of fairness and discrimination. They allow us to encode our beliefs about how we might believe that AI systems ought to behave. 

Table \ref{tab:fairness_effects} lists the normative effects of each form of fairness as a function of how the protected attribute $A=a$ affects a decision $D=d$. The total effect is the sum of the direct and indirect effects: $TE_a(d) = DE_a(d) + IE_a(d)$. By definition, the total effect must be equal to zero for it to be considered fair: $TE_a(d)=0$.

\begin{table}[htbp!]
    \caption{Normative effects of fairness}
    \begin{tabular}{c c c c c}
        \toprule
        \textbf{Effect} & \textbf{Lottery} & \makecell{\textbf{Affirmative}\\\textbf{action}} & \makecell{\textbf{Unaware-}\\\textbf{ness}} & \makecell{\textbf{Suprema-}\\\textbf{cism}}\\
        \midrule
        $DE_a(d)$ & 0 & + & 0 & --\\
        $IE_a(d)$ & -- + & -- & 0 & +\\
        $TE_a(d)$ & 0 & 0 & 0 & 0\\
        \bottomrule
    \end{tabular}
    \Description{A list of the normative effects for each form of fairness.}
    \label{tab:fairness_effects}
\end{table}

Table \ref{tab:discrimination_effects} lists the normative effects of each form of discrimination. By definition, the total effect must not be equal to zero for it to be considered discrimination: $TE_a(d) \neq 0$.

\begin{table}[htbp!]
    \caption{Normative effects of discrimination}
    \begin{tabular}{c c c c c}
        \toprule
        \textbf{Effect} & \textbf{``No''} & \textbf{Reverse} & \textbf{Covert} & \textbf{Overt}\\
        \midrule
        $DE_a(d)$ & 0 & + &  0 & --\\
        $IE_a(d)$ & + & 0 &  -- & 0\\
        $TE_a(d)$ & + & + &  -- & --\\
        \bottomrule
    \end{tabular}
    \Description{A list of the normative effects for each form of discrimination.}
    \label{tab:discrimination_effects}
\end{table}

Notice that each form of fairness is associated with a different form of discrimination, leading to four fairness and discrimination pairs. That is, what might be considered fair for one group of people may be considered biased and discriminatory for another group of people. Fairness through supremacism (Figure \ref{fig:cbn}a) and overt discrimination (Figure \ref{fig:cbn}b) are both associated with the protected attribute $A$ having a negative direct effect on the decision $D$ (\textcolor{red}{$A \rightarrow D$}). Fairness through unawareness (Figure \ref{fig:cbn}c) and covert discrimination (Figure \ref{fig:cbn}d) are both associated with the protected attribute $A$ having no direct effect on the decision $D$. Fairness through affirmative action (Figure \ref{fig:cbn}e) and reverse discrimination (Figure \ref{fig:cbn}f) are both associated with the protected attribute $A$ having a positive direct effect on the decision $D$ (\textcolor{blue}{$A \rightarrow D$}). Finally, fairness through lottery (Figure \ref{fig:cbn}g) and ``no'' discrimination (Figure \ref{fig:cbn}h) are both associated with the protected attribute $A$ having no direct effect on the decision $D$ and a positive indirect effect on the decision $D$ (\textcolor{blue}{$A \rightarrow W \rightarrow X \not\rightarrow D$}).

\subsection{The limitations of causal Bayesian networks}

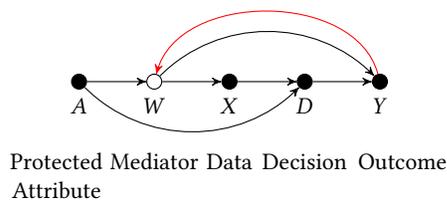
\begin{figure}[htbp!]
    \centering
    \begin{tikzpicture}[->, >=stealth', node distance=1cm, align=center]
        \node[node, label=below:$X$] (X) at (0,0) {};
        \node[node, label=below:$W$, fill=white] (W) [left of=X] {};
        \node[node, label=below:$A$] (A) [left of=W] {};
        \node[node, label=below:$D$] (D) [right of=X] {};
        \node[node, label=below:$Y$] (Y) [right of=D] {};
        \node (X2) [below of=X, yshift=-0.25cm] {Data\\};
        \node (W2) [left of=X2] {Mediator\\};
        \node (A2) [left of=W2, xshift=-0.3cm] {Protected\\Attribute};
        \node (D2) [right of=X2] {Decision\\};
        \node (Y2) [right of=D2, xshift=0.3cm] {Outcome\\};
        \path
            (A) edge node {} (W)
            (W) edge node {} (X)
            (X) edge node {} (D)
            (D) edge node {} (Y)
            (W) edge[bend left=45] node {} (Y)
            (A) edge[bend right=45] node {} (D)
            (Y) edge[bend right=75, color=red] node {} (W);
    \end{tikzpicture}
    \caption{Adding a feedback loop to the causal Bayesian network}
    \Description{A causal Bayesian network where the outcome $Y$ causes mediator $W$, creating a directed acyclic graph violation.}
    \label{fig:causal_bayesian_network_loop}
\end{figure}

CBNs have a key limitation: they can only be represented using directed acyclic graphs (DAGs). Recall from the block diagram in Figure \ref{fig:three_representations}a that outcome $Y$ could be recorded as future data $X$. We can modify the CBN from Figure \ref{fig:three_representations}b to model this feedback loop by adding a edge from outcome $Y$ to the mediator $W$ as shown in Figure \ref{fig:causal_bayesian_network_loop} (this edge has been colored red), which is then recorded as data $X$. Even though this violates the DAG constraint of CBNs, we still need to model how different fairness policies may perpetuate disparities across protected groups over time. In the next section, we will show how to overcome this limitation using causal loop diagrams.

\section{Modeling fairness as a system dynamics problem}

We can transform the block diagram from Figure \ref{fig:three_representations}a and the CBN from Figure \ref{fig:three_representations}b into a causal loop diagram (CLD) as shown in Figure \ref{fig:three_representations}c. CLDs were pioneered by Forrester \cite{forrester} and are similar to CBNs in that they allow us to symbolically encode knowledge and assumptions about cause and effect relationships between variables\cite{sterman}. But unlike CBNs, CLDs allow us to model feedback loops. This is useful for modeling why we might believe that disparities continue to exist between protected groups.

The CLD in Figure \ref{fig:three_representations}c is a simplified version of the full CLD that we will present later in this section. This shows how the same variables in the block diagram and the CBN can be mapped onto a CLD. In the center of the CLD is a node for decision $D$, which like before contains our machine learning model. This CLD is two sided, with the left side representing group A and the right side representing group B. Data from group A $X_A$ and group B $X_B$ is fed into the ML model, which renders a decision $D$ ($X_A \rightarrow D \leftarrow X_B$). Note that there is a time delay (denoted by ||) between decision $D$ and outcomes $Y_A$ and $Y_B$. The outcomes $Y_A$ and $Y_B$ are recorded as future data $X_A$ and $X_B$ ($Y_A \rightarrow X_A$ and $Y_B \rightarrow X_B$), which in turn drives a future decision $D$ ($X_A \rightarrow D \leftarrow X_B$). The outcomes $Y_A$ and $Y_B$ also influence whether the protected attribute $A_A$ or $A_B$ might be used to influence a decision $D$. For example, if group B has historically received worse outcomes that group A, then the group might demand the protected attribute $A_B$ to be affirmatively used to modify decision $D$ (fairness through affirmative action). However, the use of the protected attribute $A_A$ or $A_B$ could also impact outcomes $Y_A$ or $Y_B$.

Within a CLD, every path is labeled as having a positive (+) or negative (--) direction. If the path is positive, then changes in one variable leads to changes in another variable in the same direction. For example, an increase in one variable would lead to an increase in the other variable. Alternatively, a decrease in one variable would lead to a decrease in the other variable. If the path is negative, then changes in one variable leads to changes in another variable in the opposite direction. For example, an increase in one variable would lead to a decrease in the other variable. Alternatively, a decrease in one variable would lead to an increase in the other variable.

CLDs are represented using a combination of reinforcing loops or balancing loops. Reinforcing loops compound changes over time, creating either virtuous cycles that cause growth or vicious cycles that cause decay. Balancing loops counteract changes over time, keeping a system within a fixed state that prevents growth or decay. Reinforcing loops have an even number of negative paths, while balancing loops have an odd number of negative paths. 

We can combine reinforcing and balancing loops to create system archetypes. These represent common recurring patterns that explain counterintuitive behavior in complex systems. This concept was first coined by Senge\cite{senge} with additional research done by Meadows\cite{meadows} and Sterman\cite{sterman}. Three system archetypes seem particularly helpful for understanding the fairness problem, which we have adapted from Kim\cite{kim}.

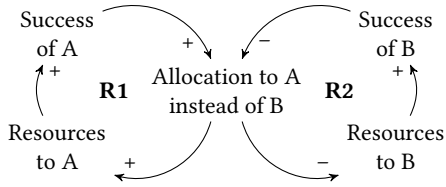
\begin{figure}[htbp!]
    \begin{tikzpicture}[->, >=stealth', align=center]
        \node (AAB) at (0,0) {Allocation to A\\instead of B};
        \node (RA) at (-2.25,-0.75) {Resources\\to A};
        \node (RB) at (2.25,-0.75) {Resources\\to B};
        \node (SA) at (-2.25,0.75) {Success\\of A};
        \node (SB) at (2.25,0.75) {Success\\of B};
        \node (R1) at (-1.5,0) {\textbf{R1}};
        \node (R2) at (1.5,0) {\textbf{R2}};
        \path (AAB) edge[bend left=45] node[very near end, above] {+} (RA);
        \path (AAB) edge[bend right=45] node[very near end, above] {--} (RB);
        \path (RA) edge[bend left=25] node[very near end, right] {+} (SA);
        \path (RB) edge[bend right=25] node[very near end, left] {+} (SB);
        \path (SA) edge[bend left=45] node[very near end, left] {+} (AAB);
        \path (SB) edge[bend right=45] node[very near end, right] {--} (AAB);
    \end{tikzpicture}
    \caption{Success to the successful archetype}
    \Description{Modeling fairness using the success to the successful archetype.}
    \label{fig:success}
\end{figure}

The first archetype that we will examine is the \textbf{``success to the successful''} archetype, which is shown in Figure \ref{fig:success}. This archetype consists of two reinforcing loops, R1 and R2. Let's say that we have two equally capable groups that are competing for the same resources: groups A and B. If group A was initially given more resources, then that group will become more successful. That success provides justification to allocate even more resources to group A. This is shown by reinforcing loop R1. However, that could mean fewer resources are allocated to group B, which causes that group to become less successful. This results in fewer resources being allocated to group B. This is shown by reinforcing loop R2. Over the long run, R1 could create a virtuous cycle of growth for group A, while R2 could create a vicious cycle of decay for group B.

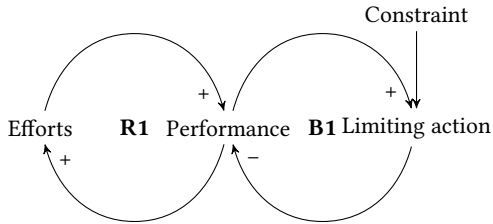
\begin{figure}[htbp!]
    \begin{tikzpicture}[->, >=stealth', align=center]
        \node (P) at (0,0) {Performance};
        \node (E) at (-2.5,0) {Efforts};
        \node (LA) at (2.5,0) {Limiting action};
        \node (C) at (2.5,1.5) {Constraint};
        \node (R1) at (-1.25,0) {\textbf{R1}};
        \node (B1) at (1.25,0) {\textbf{B1}};
        \draw (P) to[out=-105,in=0] (-1.25,-1.25) to[out=180,in=-75] node[very near end, right] {+} (E);
        \draw (E) to[out=75,in=180] (-1.25,1.25) to[out=0,in=105] node[very near end, left] {+} (P);
        \draw (P) to[out=75,in=180] (1.25,1.25) to[out=0,in=105] node[very near end, left] {+} (LA);
        \draw (LA) to[out=-105,in=0] (1.25,-1.25) to[out=180,in=-75] node[very near end, right] {--} (P);
        \path (C) edge node[very near end, right] {} (LA);
    \end{tikzpicture}
    \caption{Limits to success archetype}
    \Description{Modeling fairness using the limits to success archetype.}
    \label{fig:limits}
\end{figure}

The next archetype that we will examine is the \textbf{``limits to success''} archetype, which is shown in Figure \ref{fig:limits}. This archetype consists of a reinforcing loop R1 and a balancing loop B1. Let's say that a system's performance continues to increase as a result of past efforts. This is shown by the reinforcing loop R1. But as performance increases, it begins to plateau as it reaches some limiting action. The limiting action is governed by some constraint that prevents further performance improvement. This is shown by the balancing loop B1. Even as efforts increase, the limiting action acts as a ceiling on performance.

\begin{figure}[htbp!]
    \begin{tikzpicture}[->, >=stealth', align=center]
        \node (PS) at (0,0) {Problem\\symptom};
        \node (IS) at (0,-2) {Internal\\solution};
        \node (EI) at (0,2) {External\\intervention};
        \node (DEI) at (2.5,0) {Dependence\\on external\\intervention};
        \node (B1) at (0,-1) {\textbf{B1}};
        \node (B2) at (0,1) {\textbf{B2}};
        \node (R1) at (1.2,0) {\textbf{R1}};
        \path (PS) edge[bend left=45] node[very near end, left] {+} (IS);
        \path (IS) edge[bend left=45] node[very near end, right] {--} (PS);
        \path (PS) edge[bend left=45] node[very near end, right] {+} (EI);
        \path (EI) edge[bend left=45] node[very near end, left] {--} (PS);
        \path (EI) edge[bend left=30] node[very near end, left] {+} (DEI);
        \path (DEI) edge[bend left=30] node[very near end, above] {--} (IS);
        \node[rotate=90] at (-0.675,-1) {\LARGE ||};
    \end{tikzpicture}
    \caption{Shifting the burden / addiction archetype}
    \Description{Modeling fairness using the success to the successful archetype.}
    \label{fig:shifting}
\end{figure}
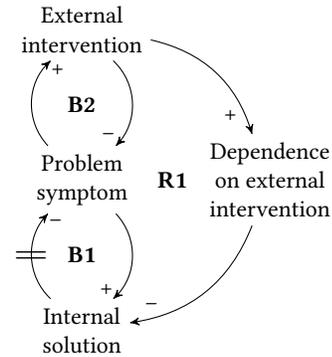

The final system archetype that we will examine is the \textbf{``shifting the burden / addiction''} archetype, which is shown in Figure \ref{fig:shifting}. Let's say that there is a problem symptom and two choices for a solution: an internal solution and an external intervention. The internal solution takes a long time to correct for the problem symptom (the double hash marks || between the internal solution and the problem symptom represent a time delay), meaning that the problem symptom would continue to persist in the short term. On the other hand, the external intervention might alleviate the problem symptom immediately. But an unintended side effect of this solution is that it could create a dependence on the external intervention, which diverts attention away from an internal solution. This means that the problem symptom would actually persist over the long term, offsetting the benefit of the external intervention. This is called the shifting the burden archetype because the burden of solving the problem symptom is shifted from one party to another. It is also called an addiction archetype because people who rely on an external intervention may become addicted to that solution over the long term.

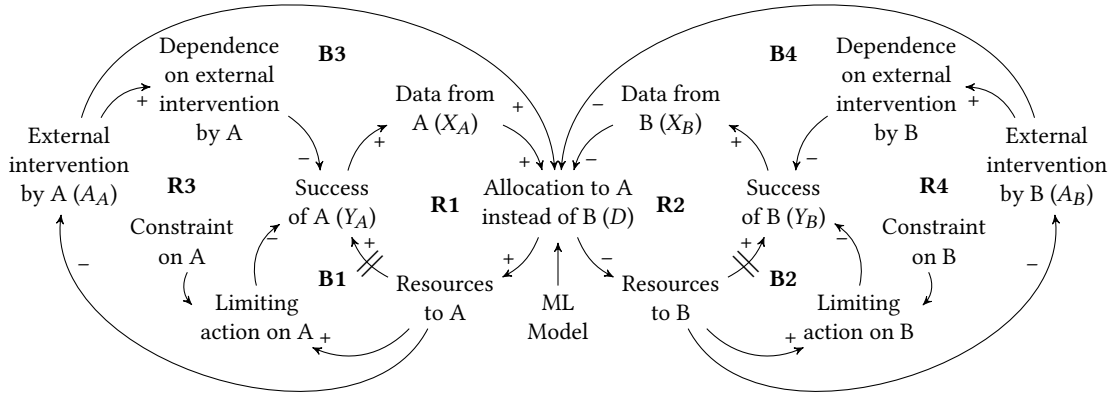
\begin{figure*}[htbp!]
    \begin{tikzpicture}[->, >=stealth', align=center]
        \node (AAB) at (0,0) {Allocation to A\\instead of B ($D$)};
        \node (RA) at (-1.5,-1.25) {Resources\\to A};
        \node (RB) at (1.5,-1.25) {Resources\\to B};
        \node (SA) at (-3,0) {Success\\of A ($Y_A$)};
        \node (SB) at (3,0) {Success\\of B ($Y_B$)};
        \node (DA) at (-1.5,1.25) {Data from\\A ($X_A$)};
        \node (DB) at (1.5,1.25) {Data from\\B ($X_B$)};
        \node (R1) at (-1.5,0) {\textbf{R1}};
        \node (R2) at (1.5,0) {\textbf{R2}};
        \path (AAB) edge[bend left=20, color=red] node[very near end, above] {+} (RA);
        \path (AAB) edge[bend right=20, color=blue] node[very near end, above] {--} (RB);
        \path (RA) edge[bend left=20, color=red] node[very near end, right] {+} (SA);
        \path (RB) edge[bend right=20, color=red] node[very near end, left] {+} (SB);
        \path (SA) edge[bend left=25, color=red] node[very near end, below] {+} (DA);
        \path (SB) edge[bend right=25, color=red] node[very near end, below] {+} (DB);
        \path (DA) edge[bend left=25, color=red] node[very near end, left] {+} (AAB);
        \path (DB) edge[bend right=25, color=blue] node[very near end, right] {--} (AAB);
        \node (LAA) at (-4,-1.5) {Limiting\\action on A};
        \node (LAB) at (4,-1.5) {Limiting\\action on B};
        \node (B1) at (-3,-1) {\textbf{B1}};
        \node (B2) at (3,-1) {\textbf{B2}};
        \path (RA) edge[bend left=35, color=red] node[very near end, above] {+} (LAA);
        \path (RB) edge[bend right=35, color=red] node[very near end, above] {+} (LAB);
        \path (LAA) edge[bend left=35, color=blue] node[very near end, below] {--} (SA);
        \path (LAB) edge[bend right=35, color=blue] node[very near end, below] {--} (SB);
        \node (CA) at (-5,-0.5) {Constraint\\on A};
        \node (CB) at (5,-0.5) {Constraint\\on B};
        \path (CA) edge[bend right=35] node[midway, right] {} (LAA);
        \path (CB) edge[bend left=35] node[midway, left] {} (LAB);
        \node (EIA) at (-6.5,0.5) {External\\intervention\\by A ($A_A$)};
        \node (EIB) at (6.5,0.5) {External\\intervention\\by B ($A_B$)};
        \path (RA) edge[bend left=80, color=blue] node[very near end, right] {--} (EIA);
        \path (RB) edge[bend right=80, color=blue] node[very near end, left] {--} (EIB);
        \path (EIA) edge[bend left=80, color=red] node[very near end, left] {+} (AAB);
        \path (EIB) edge[bend right=80, color=blue] node[very near end, right] {--} (AAB);
        \node (B3) at (-3,2) {\textbf{B3}};
        \node (B4) at (3,2) {\textbf{B4}};
        \node (DEIA) at (-4.5,1.5) {Dependence\\on external\\intervention\\by A};
        \node (DEIB) at (4.5,1.5) {Dependence\\on external\\intervention\\by B};
        \path (EIA) edge[bend left=30, color=red] node[very near end, below] {+} (DEIA);
        \path (EIB) edge[bend right=30, color=red] node[very near end, below] {+} (DEIB);
        \path (DEIA) edge[bend left=25, color=blue] node[very near end, left] {--} (SA);
        \path (DEIB) edge[bend right=25, color=blue] node[very near end, right] {--} (SB);
        \node (R3) at (-5,0.25) {\textbf{R3}};
        \node (R4) at (5,0.25) {\textbf{R4}};
        \node (ML) at (0,-1.5) {ML\\Model};
        \draw[->, shorten >=2pt] (ML) -- (AAB);
        \node[rotate=-45, color=red] at (-2.5,-0.8) {\LARGE ||};
        \node[rotate=45, color=red] at (2.5,-0.8) {\LARGE ||};
        \end{tikzpicture}
    \caption{Causal loop diagram representation of algorithmic fairness}
    \Description{Modeling fairness using the success to the successful, limits to success, and the shifting the burden / addiction archetypes.}
    \label{fig:archetypes}
\end{figure*}

In Figure \ref{fig:archetypes}, we combine these three archetypes into a single CLD to model the algorithmic fairness problem. This diagram is an expanded version of the CLD in Figure \ref{fig:three_representations}c. We begin with the ``success to the successful'' archetype. A machine learning model is embedded inside decision $D$, which makes allocation decisions based on decision $X$ and has a tendency to favor one group over another. Those decisions lead to resources being assigned to each group, which eventually determines that group's success and is measured as an outcome $Y$. Past success $Y$ becomes future data $X$, which is then fed back into the machine learning model to drive a future allocation decision $D$. This is represented using reinforcing loops R1 and R2. Since the allocation decisions $D$ are initially only driven by data $X$, success to the successful is equivalent to fairness through unawareness (Figure \ref{fig:cbn}c).

We next add in the ``limits to success'' archetype. As resources to group A or group B increases, it may hit up against a limiting action that is governed by come constraint. This then reduces the success of that group as measured by outcome $Y$. As the group's success becomes limited, this is reflected in the data $X$ for that group. This leads to a reduction in the allocation decision $D$ of resources to the previously successful group. This is represented by balancing loops B1 and B2. Whereas success to the successful causes a previously successful group to become more successful, limits to success may give a less successful group a chance to catch up.

Finally, we add in the ``shifting the burden / addiction'' archetype. If one group is allocated less resources than another group, then it could breed resentment by the disadvantaged group. That group could then demand that the protected attribute $A$ be used to externally intervene on the allocation decision $D$. This external intervention is equivalent to fairness through affirmative action (Figure \ref{fig:cbn}e). While it might reduce allocation disparities in the short term, it may actually create a dependence on the external intervention over the long term. This is because if allocation decisions $D$ are not based purely on data $X$, which is a measure of mediator $W$ (e.g. merit, qualifications), then the assignment of resources to the disadvantaged group may not lead to their success $Y$. For example, approving a loan to a minority applicant even though the applicant is not creditworthy could lead to a worse outcome (default) than if the loan was denied in the first place. That decreased success in outcome $Y$ for the less successful group could trigger a demand for an even stronger external intervention $A$ on the decision $D$, reinforcing this problem. In the meanwhile, attention may be taken away from the internal solution of optimizing for the mediator $W$ and the data $X$ so that the disadvantaged group would perform better under fairness through unawareness. This may especially be the case if the disadvantaged group believes that the mediator $W$ is biased and therefore does not focus on optimizing that metric.

\subsection{Positive effects of fairness and discrimination} \label{positive}

Whereas CBNs are very effective at modeling the normative effects of fairness and discrimination, CLDs are very effective at modeling the positive effects. Recall from Hume's law that positive effects model what is, while normative effects model what ought to be (i.e. the is--ought problem).

Table \ref{tab:positive_effects} lists the positive effects of the different forms of fairness and discrimination. As mentioned before, each form of fairness is associated with a form of discrimination, leading to the same direct effect $DE_a(d)$. Whereas normative effects differ based on our prior beliefs about where bias exists in a model, positive effects do not take into account our prior beliefs of bias. However our analysis of positive effects depends on whether we analyze disparities across groups from a short-term or a long-term perspective. Notice that for fairness through supremacism or overt discrimination, since a disadvantaged group is being overtly disadvantaged, there is a doubly negative total effect ($TE_a(d)\ll0$).

\begin{table}[htbp!]
    \caption{Positive effects of fairness and discrimination}
    \begin{tabular}{c c c c c}
        \toprule
        \textbf{Fairness} & \textbf{Lottery} & \makecell{\textbf{Affirmative}\\\textbf{action}} & \makecell{\textbf{Unaware-}\\\textbf{ness}} & \makecell{\textbf{Suprema-}\\\textbf{cism}}\\
        \midrule
        \makecell{\textbf{Discrim-}\\\textbf{ination}} & \textbf{``No''} & \textbf{Reverse} & \textbf{Covert} & \textbf{Overt}\\
        \midrule
        \multicolumn{5}{c}{\textit{Short-term}}\\
        \midrule
        $DE_a(d)$ & 0 & + &  0 & --\\
        $IE_a(d)$ & 0 & -- &  -- & --\\
        $TE_a(d)$ & 0 & 0 &  -- & -- --\\
        \midrule
        \multicolumn{5}{c}{\textit{Long-term}}\\
        \midrule
        $DE_a(d)$ & 0 & + &  0 & --\\
        $IE_a(d)$ & 0 & -- -- &  -- + & --\\
        $TE_a(d)$ & 0 & -- &  0 & -- --\\
        \bottomrule
    \end{tabular}
    \Description{A list of the positive effects for each form of discrimination.}
    \label{tab:positive_effects}
\end{table}

During the short-term, we only examine the ``success to the successful'' archetype as shown in Figure \ref{fig:archetypes} through feedback loops R1 and R2. A policy of fairness through unawareness would lead to a disadvantaged group receiving worse decisions than an advantaged group ($TE_a(d)<0$). This is because fairness through unawareness would replicate and reinforce the past success of the advantaged group over the disadvantaged group ($IE_a(d)<0$). But an affirmative action policy could help to equalize the decision rates across both groups. Therefore, the total effect from fairness through unawareness is negative ($TE_a(d)<0$) and the total effect from affirmative action is zero ($TE_a(d)=0$).

During the long-term, we add in the effects of the ``limits to success'' archetype as shown in Figure \ref{fig:archetypes} through feedback loops B1 and B2. This archetype puts a constraint on how successful the advantaged group can become, reducing the disparities between the advantaged group and the disadvantaged group ($IE_a(d)>0$). We also add in the effects of the ``shifting the burden / addiction'' archetype as shown in Figure \ref{fig:archetypes} through feedback loops R3, B3, R4, and B4. This archetype can create a dependency by the disadvantaged group on the external intervention and harm their long-term success ($IE_a(d)<0$). Therefore, the  total effect from fairness through unawareness is zero ($TE_a(d)=0$) and the total effect from affirmative action is negative ($TE_a(d)<0$). That is, the total effects between the short-term and long-term versions of these fairness policies may flip direction.

\section{Modeling fairness as a causal hierarchy} \label{hierarchy}

Systems thinking allows us to combine three previously independent analytical techniques: machine learning, causal inference, and system dynamics. Doing so allows us to capture the emergent behavior from three different aspects of the fairness problem. We can think of CBNs as a higher level representation of supervised machine learning models, which emerges when we ``think outside of the black box.'' We can think of CLDs as a higher level representation of CBNs, which emerges when we violate the directed acyclic graph constraint and create a feedback loop. Another way to think of this is that a CBN may act as a ``wrapper'' around a supervised ML model and as a ``snapshot'' of a CLD.

\begin{figure*}[htbp!]
    \begin{tikzpicture}[->, >=stealth', align=left]
        \node[wide_block, text width=7.75cm] (P1) at (0,0) {\textbf{Layer 1. Association}\\\textbf{Activity:} Seeing, observing\\\textbf{Question:} What if I see...?\\\textbf{Example:} What does a symptom tell me about a disease?};
        \node[wide_block, text width=7.75cm] (P2) at (0,2.25) {\textbf{Layer 2. Intervention}\\\textbf{Activity:} Doing, intervening\\\textbf{Question:} What if I do...? How?\\\textbf{Example:} If I take aspirin, will my headache be cured?};
        \node[wide_block, text width=7.75cm] (P3) at (0,4.5) {\textbf{Layer 3. Counterfactuals}\\\textbf{Activity:} Imagining, retrospection, understanding\\\textbf{Question:} What if I had done...? Why?\\\textbf{Example:} Was it the aspirin that stopped my headache?};
        \node[wide_block, text width=6cm] (L1) at (8,0) {\textbf{Layer 1. Association}\\\textbf{Field:} Supervised machine learning\\\textbf{Representation:} Block diagrams\\\textbf{Application:} Prediction};
        \node[wide_block, text width=6cm] (L2) at (8,2.25) {\textbf{Layer 2. Static causation}\\\textbf{Field:} Causal inference (Judea Pearl)\\\textbf{Representation:} Causal Bayesian networks\\\textbf{Application:} Control};
        \node[wide_block, text width=6cm] (L3) at (8,4.5) {\textbf{Layer 3. Dynamic causation}\\\textbf{Field:} System dynamics (Jay Forrester)\\\textbf{Representation:} Causal loop diagrams\\\textbf{Application:} Simulation};
        \node (L12) at (8, 1.125) {\textcolor{red}{$\uparrow$ \textbf{Think outside the black box!} $\uparrow$}};
        \node (L23) at (8, 3.375) {\textcolor{red}{$\uparrow$ \textbf{Violate the DAG constraint!} $\uparrow$}};
        \draw (P1.east) -- (L1.west);
        \draw (P2.east) -- (L2.west);
        \draw (P3.east) -- (L2.west);
    \end{tikzpicture}
    \caption{Causal hierarchy representation of algorithmic fairness}
    \Description{Adding a system dynamics layer to Pearl's causal hierarchy.}
    \label{fig:hierarchy}
\end{figure*}
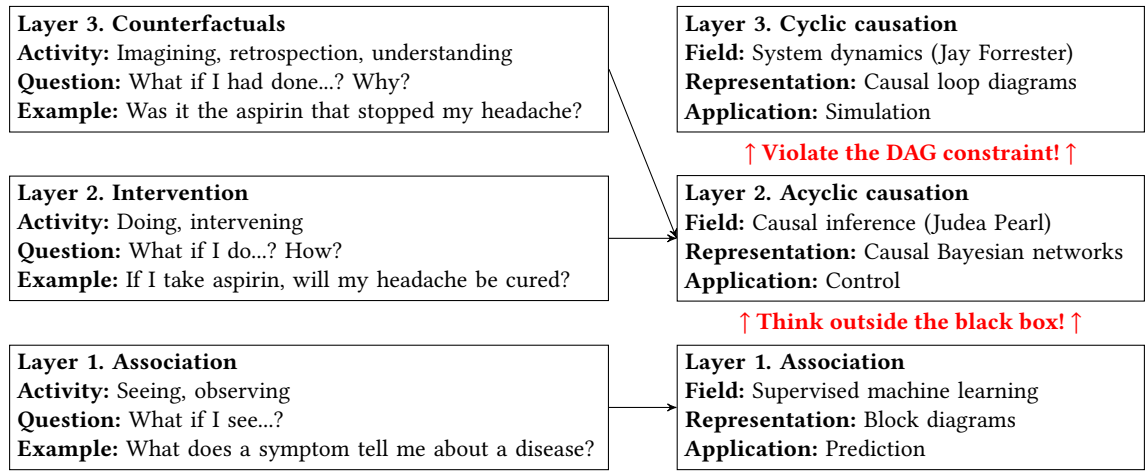

However, this suggests the need for a new causal hierarchy to model the algorithmic fairness problem. The left side of Figure \ref{fig:hierarchy} shows the causal hierarchy proposed by Pearl, which is not optimized for supervised machine learning and does not incorporate system dynamics. The right side of Figure \ref{fig:hierarchy} is a proposed new causal hierarchy that more explicitly models the emergent behavior between supervised machine learning, causal inference, and system dynamics.

We can directly map layer 1 of Pearl's causal hierarchy (association) to layer 1 of this new causal hierarchy (association), but we explicitly define this layer to be used for supervised machine learning. We can then combine layers 2 and 3 of Pearl's causal hierarchy (intervention and counterfactuals) into layer 2 of this new causal hierarchy (acyclic causation). Finally, we add a new layer 3 to the causal hierarchy for system dynamics (cyclic causation).

Each of the layers in this new causal hierarchy have unique properties and behaviors that make them optimized for different applications. The first layer is associated with supervised machine learning and is optimized for prediction. The second layer is associated with causal inference and is optimized for control. The third layer is associated with system dynamics and is optimized for simulation. These three new causal layers make up an expanded causal hierarchy that is consistent with systems thinking principles.

\section{Modeling fairness as a systems map} \label{systems_map}

\begin{figure*}[htbp!]
    \begin{tikzpicture}[<->, >=stealth', align=left]]
        \node[block] (S) at (0,4) {\LARGE \textbf{Social}};
        \node[block] (T) at (9.5,4) {\LARGE \textbf{Technical}};
        \node[block, draw] (P) at (0,3) {\textbf{Politics}};
        \node[small_block, align=left] at (0,2.6) {\textcolor{blue}{Left-wing}\hfill\textcolor{red}{Right-wing}};
        \node[block, draw] (So) at (3.5,1) {\textbf{Sociology}};
        \node[small_block, align=left] at (3.5,0.6) {\textcolor{blue}{Structure}\hfill\textcolor{red}{Agency}};
        \node[block, draw] (Ps) at (3.5,-1) {\textbf{Psychology}};
        \node[small_block, align=left] at (3.5, -1.4) {\textcolor{blue}{External}\hfill\textcolor{red}{Internal}\\\textcolor{blue}{attribution}\hfill\textcolor{red}{attribution}};
        \node[block, draw] (A) at (0,-3) {\textbf{Anthropology}};
        \node[small_block, align=left] at (0,-3.4) {\textcolor{blue}{Social}\hfill\textcolor{red}{Cultural}};
        \node[block, draw] (E) at (-3.5,-1) {\textbf{Economics}};
        \node[small_block, align=left] at (-3.5,-1.4) {\textcolor{blue}{Socialism}\hfill\textcolor{red}{Capitalism}};
        \node[block, draw] (L) at (-3.5,1) {\textbf{Law}};
        \node[small_block, align=left] at (-3.5,0.6) {\textcolor{blue}{Disparate}\hfill\textcolor{red}{Disparate}\\\textcolor{blue}{impact}\hfill\textcolor{red}{treatment}};
        \node[block, draw] (SD) at (9.5,3) {\textbf{System Dynamics}};
        \node[small_block, align=left] at (9.5,2.6) {\textcolor{blue}{Success}\hfill\textcolor{red}{Shifting the}\\\textcolor{blue}{to the}\hfill\textcolor{red}{burden /}\\\textcolor{blue}{successful}\hfill\textcolor{red}{Addiction}};
        \node[block, draw] (CI) at (9.5,0) {\textbf{Causal Inference}};
        \node[small_block, align=left] at (9.5,-0.4) {\textcolor{blue}{First law:}\hfill\textcolor{red}{Second law:}\\\textcolor{blue}{Counterfactuals}\hfill\textcolor{red}{Conditional}\\\textcolor{blue}{and}\hfill\textcolor{red}{independence}\\\textcolor{blue}{interventions}};
        \node[block, draw] (ML) at (9.5,-3) {\textbf{Machine Learning}};
        \node[small_block, align=left] at (9.5,-3.4) {\textcolor{blue}{Fairness}\hfill\textcolor{red}{Fairness}\\\textcolor{blue}{through}\hfill\textcolor{red}{through}\\\textcolor{blue}{affirmative}\hfill\textcolor{red}{unawareness}\\\textcolor{blue}{action}};

        \draw[-, dashed] (6.5,4.25) to (6.5,-5);
        \node[fill=white] (CB) at (6.5,0) {\textbf{Causal}\\\textbf{Bridge}};

        \draw (0,0) to (P.south);
        \draw (0,0) to (So.west);
        \draw (0,0) to (Ps.west);
        \draw (0,0) to (A.north);
        \draw (0,0) to (E.east);
        \draw (0,0) to (L.east);
        \draw (0,0) to (CI.west);
        \draw (CI.north) to (SD.south);
        \draw (CI.south) to (ML.north);

        \node at (0, 0) {\includegraphics[scale=0.025]{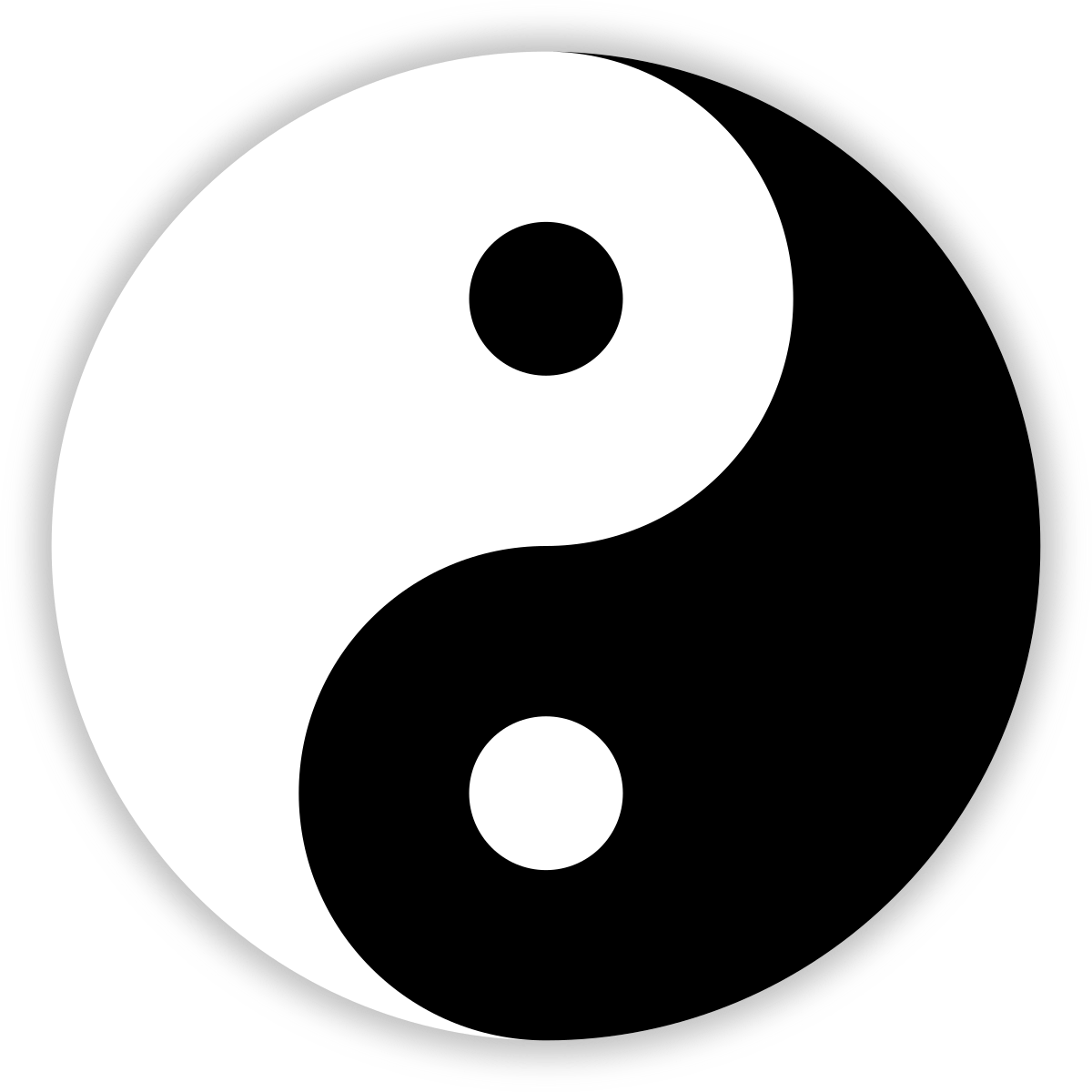}};
    \end{tikzpicture}
    \caption{Systems map representation of algorithmic fairness}
    \Description{Systems map showing the causal links between the social sciences and computer science.}
    \label{fig:system_map}
\end{figure*}

We can create an even higher level representation of the algorithmic fairness problem called a systems map. This is shown in Figure \ref{fig:system_map}. On the left side of the systems map, we use a connection circle to show the different fields in the social sciences that are related to fairness. These ideas were described in greater detail in Section \ref{theory}. On the right side of the systems map, we use a causal hierarchy to show the different technical aspects of the fairness problem. These ideas were described in greater detail in Section \ref{hierarchy}.

As mentioned before, algorithmic fairness is a sociotechnical problem. We need to find a way to bridge the social and technical aspects of the fairness problem. In each of the social science fields, there exists a dichotomy that represents the two sides of the fairness problem. This allows us to understand the fundamentally ideological and political debate over fairness, which centers around how we attribute the causes for disparities across protected groups. The systems map shows a causal bridge that links this dualism in the social sciences to Pearl's two fundamental laws of causal inference. This systems map provides a very compact and high level representation for how the different fields of fairness may relate to each other.

\section{Conclusion}

Systems thinking provides us with a novel way to model the algorithmic fairness problem by ``thinking outside the black box.'' While a large amount of AI research has been focused on the hidden layers that are ``inside'' a black box model, this paper has focused on the causal layers that are ``outside'' a black box model. These outer layers are where we can encode our knowledge and assumptions about where we might believe that bias occurs in the data generating process. We showed that the first outer layer can be represented using causal Bayesian networks and the second outer layer can be represented using causal loop diagrams. This gives us an elegant way of modeling the emergent behavior between machine learning, causal inference, and system dynamics.

This approach allow us to develop a more unified and holistic view of the fairness and machine learning problem, providing essential connections between computer science, politics, and the law. This is the key to building AI systems that are aligned with society's shared democratic values and the policymakers' political agendas. It will enable businesses and regulators to communicate with each other using a common causal language, one that can be used by both data scientists and lawyers. These models may help to identify mathematical flaws in the existing laws and regulations. They could help businesses to comply with antidiscrimination laws and regulations, reducing business uncertainty and expensive litigation while providing a much needed layer of transparency and trust for consumers of AI products and services.

Systems thinking supplies the missing mathematical models that are needed to build responsible AI systems. To solve fairness, we need to bring together both the social and technical aspects of what is ultimately a complex systems problem. On the social side, it requires modeling the diversity of perspectives that come from both sides of the political spectrum. On the technical side, it requires unifying analytical techniques that were previously thought of as being independent. By unifying all of these concepts together, we hope that this enables the development of AI policies and systems that could lead to pragmatic, real-world solutions for reducing disparities across protected groups in society.

\bibliographystyle{ACM-Reference-Format}
\bibliography{facct}

\end{document}